\documentclass[letterpaper, 10 pt, conference]{ieeeconf}  

\IEEEoverridecommandlockouts                              

\usepackage{graphicx}      
\usepackage{mathptmx}      
\usepackage{amsmath}       
\usepackage{amssymb}       
\usepackage[utf8]{inputenc} 

\usepackage{booktabs}      
\usepackage{tabularx}      

\usepackage{algorithm}
\usepackage{algpseudocode}

\usepackage{cite}
\usepackage{url}           

\usepackage[hidelinks]{hyperref} 

\title{\LARGE \bf
EvoGym\underline{CM}: Harnessing \underline{C}ontinuous \underline{M}aterial Stiffness for Soft Robot Co-Design
}

\author{Le Shen$^{1}$, Kangyao Huang$^{2}$, Wentao Zhao$^{2}$, and Huaping Liu$^{2*}$
\thanks{$^{*}$ Corresponding author: {\tt hpliu@tsinghua.edu.cn}}
\thanks{$^{1}$ College of Control Science and Engineering, Zhejiang University.}%
\thanks{$^{2}$ Department of Computer Science and Technology, Tsinghua University.}%
\thanks{Source code is available at: 
\url{https://github.com/coke-2004/EvoGymCM}} 
}

\begin{document}

\maketitle
\thispagestyle{empty}
\pagestyle{empty}

\begin{abstract}

In the automated co-design of soft robots, precisely adapting the material stiffness field to task environments is crucial for unlocking their full physical potential. However, mainstream platforms (e.g., EvoGym) strictly discretize the material dimension, artificially restricting the design space and performance of soft robots. To address this, we propose EvoGymCM \textit{(EvoGym with \underline{C}ontinuous \underline{M}aterials)}, a benchmark suite formally establishing continuous material stiffness as a first-class design variable alongside morphology and control. Aligning with real-world material mechanisms, EvoGymCM introduces two settings: (i) \textit{EvoGymCM-R (Reactive)}, motivated by programmable materials with dynamically tunable stiffness; and (ii) \textit{EvoGymCM-I (Invariant)}, motivated by traditional materials with invariant stiffness fields. To tackle the resulting high-dimensional coupling, we formulate two Morphology--Material--Control co-design paradigms: (i) \textit{Reactive-Material Co-Design}, which learns real-time stiffness tuning policies to guide programmable materials; and (ii) \textit{Invariant-Material Co-Design}, which jointly optimizes morphology and fixed material fields to guide traditional material fabrication. Systematic experiments across diverse tasks demonstrate that continuous material optimization boosts performance and unlocks synergy across morphology, material, and control.

\end{abstract}

\section{Introduction}

\begin{figure*}[t] 
    \centering
    \includegraphics[width=1.0\textwidth]{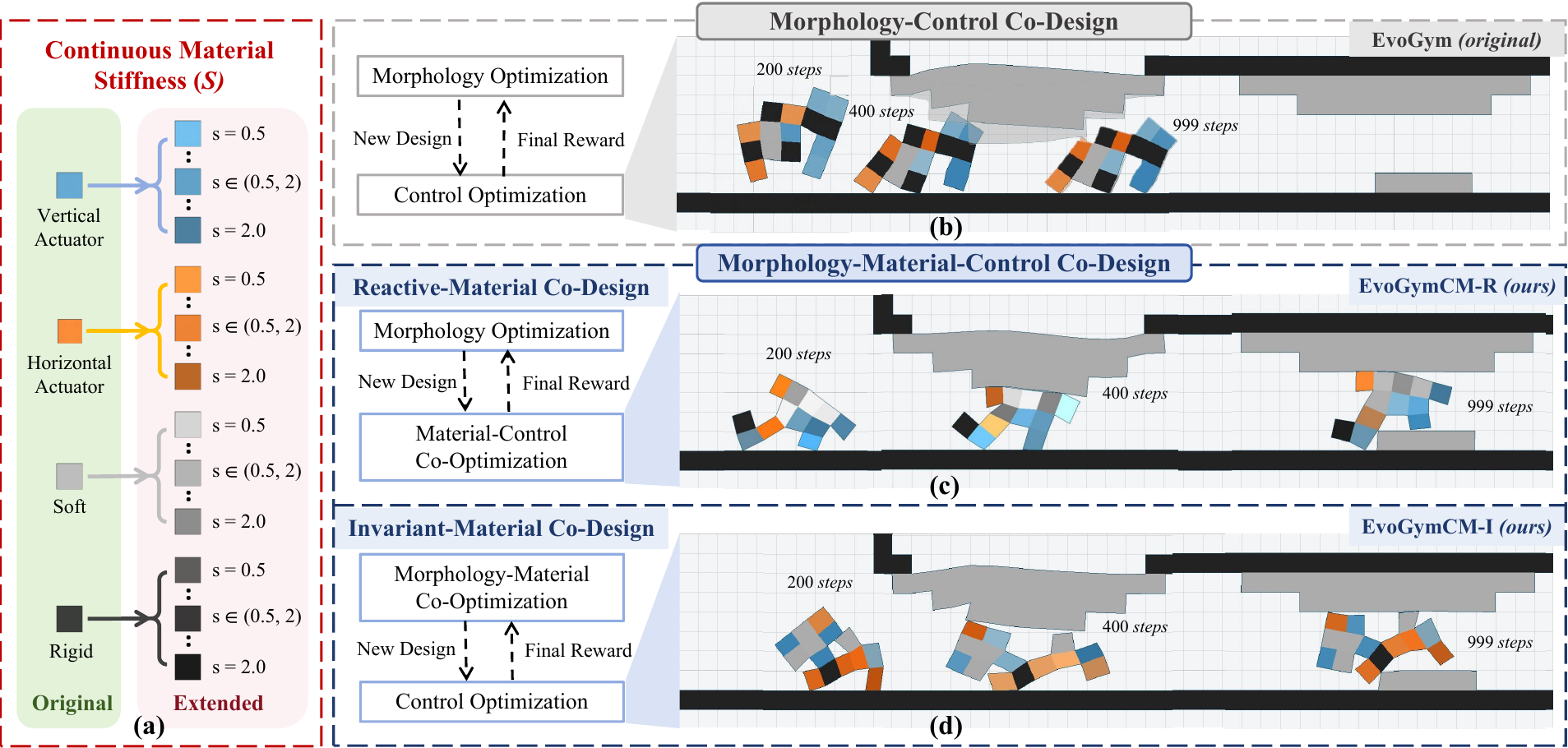} 
    \caption{\textbf{Overview of EvoGymCM and the proposed co-design paradigms.} \textbf{(a)} EvoGymCM introduces continuous material stiffness, overcoming the discrete limitations of \textbf{(b)} standard EvoGym. This continuous formulation enables two distinct paradigms: \textbf{(c)} Reactive-Material Co-Design drives real-time stiffness tuning in the inner loop; whereas \textbf{(d)} Invariant-Material Co-Design optimizes stiffness configurations in the outer loop, keeping the material field strictly invariant during evaluation.}
    \label{fig:main}
\end{figure*}

Automated design is replacing manual approaches as the pivotal paradigm to develop high-performance soft robots\cite{cheney_unshackling_2014, howard_evolving_2019}. While their intrinsic compliance offers unique advantages in unstructured environments\cite{trivedi2008soft}, it also introduces infinite degrees of freedom and highly nonlinear dynamics. This complexity renders traditional trial-and-error design labor-intensive and sub-optimal. Consequently, algorithm-driven optimization within simulation environments has become indispensable to navigate this vast design space\cite{sims_evolving_1994, liu2025embodied}.

To unlock the full physical potential of soft robots, optimal design requires precisely tailoring the material stiffness field to the task environment\cite{rus_design_2015, manti2016stiffening}. The stiffness distribution directly governs the system's deformation patterns and force transmission. Even under fixed morphology and control policies, reconfiguring the material field can yield drastically different dynamic behaviors. Therefore, elevating material stiffness from a passive attribute to a first-class design variable is crucial to overcome existing performance bottlenecks.

However, mainstream automated co-design platforms (e.g., EvoGym\cite{bhatia_evolution_2021}) suffer from a fundamental bottleneck in material representation, artificially restricting the design space
and performance of soft robots. For instance, while EvoGym has established a standardized evaluation framework for morphology-control co-design, it possesses an inherent limitation: robots are composed of a finite set of discrete voxel types (e.g., soft, rigid, and actuated) with strictly fixed stiffness properties. This paradigm reduces the natural continuum of physical materials to a discrete combinatorial selection. Consequently, this coarse material resolution constrains the design space and prevents soft robots from unlocking their full potential.

To bridge this critical gap, we introduce the continuous material stiffness dimension into EvoGym—a non-trivial expansion that gives rise to the EvoGymCM (\textit{EvoGym with \underline{C}ontinuous \underline{M}aterial}) benchmark suite. As illustrated in Fig.~\ref{fig:main}(a), unlike standard EvoGym's strict discretization of material stiffness, this novel suite formally establishes it as a first-class design variable alongside morphology and control. This integration drives a fundamental paradigm shift from discrete combinatorial search to continuous field optimization. Consequently, EvoGymCM exponentially expands the design space, unlocking unprecedented physical capabilities for soft robots.

Furthermore, to align with real-world physical mechanisms, EvoGymCM introduces two benchmark settings: 
(i) \textbf{EvoGymCM-R (Reactive)}, motivated by programmable materials: the material stiffness can be dynamically tuned during environmental interactions; and 
(ii) \textbf{EvoGymCM-I (Invariant)}, motivated by traditional materials: the optimized stiffness field remains invariant post-fabrication.

To tackle the formidable challenge of high-dimensional and non-linear coupling across morphology, material, and control, we integrate these three facets into a unified bi-level optimization framework\cite{sinha2017review}. By structuring the problem into an outer loop for physical design and an inner loop for dynamic evaluation, we formulate two tailored Morphology--Material--Control co-design paradigms:

\begin{itemize}
    \item[(i)] \textbf{Reactive-Material Co-Design}: As shown in Fig.~\ref{fig:main}(c), this paradigm, tailored for \textit{EvoGymCM-R}, pairs outer-loop morphology evolution with an inner-loop Material--Control Co-Optimization module, learning real-time stiffness tuning policies during environmental interactions to guide programmable materials.
    
    \item[(ii)] \textbf{Invariant-Material Co-Design}: As shown in Fig.~\ref{fig:main}(d), this paradigm, tailored for \textit{EvoGymCM-I}, jointly searches for optimal morphology and material stiffness fields in the outer loop, providing quantitative guidance for traditional material fabrication parameters (e.g., mixing ratios and infill densities).
\end{itemize}

The main contributions are summarized as follows:
\begin{itemize}
    \item \textbf{The EvoGymCM Benchmark Suite.} We establish continuous material stiffness as a first-class design variable alongside morphology and control, and introduce two settings: (i) \textit{EvoGymCM-R (Reactive)} for dynamically programmable materials, and (ii) \textit{EvoGymCM-I (Invariant)} for traditional fixed-stiffness materials.
    
    \item \textbf{Morphology--Material--Control Co-Design.} We propose two tailored paradigms: (i) \textit{Reactive-Material Co-Design} for learning real-time stiffness-tuning policies to guide programmable materials, and (ii) \textit{Invariant-Material Co-Design} for optimizing static stiffness fields to guide physical fabrication.
    
    \item \textbf{Systematic Validation.} We conduct extensive experiments, demonstrating that continuous material optimization boosts performance and unlocks synergy across morphology, material, and control.
\end{itemize}

The remainder of this paper is organized as follows. Section II reviews related work. Sections III and IV detail the EvoGymCM benchmark suite and the proposed co-design algorithmic paradigms, respectively. Section V presents the experimental results, followed by conclusions in Section VI.

\section{Related Work}

\subsection{Morphology and Control Co-Design in Soft Robotics} 

The co-design of morphology and control forms a fundamental paradigm in soft robotics\cite{huang2024competevo}. Early explorations in this domain successfully co-evolved the morphologies and neural controllers of virtual creatures \cite{sims_evolving_1994}. Driven by the diverse capabilities of continuous deformations, this paradigm naturally expanded into soft robotics. Initial efforts in automated soft robot design focused on voxel-based structures \cite{hiller_automatic_2012}, which were later enhanced through generative encodings like CPPN-NEAT \cite{cheney_unshackling_2014}. However, the reliance on customized physics engines initially hindered standardized comparisons across different methods. This bottleneck was ultimately resolved by the introduction of Evolution Gym (EvoGym), a large-scale benchmark specifically tailored for the co-optimization of soft robot morphology and control \cite{bhatia_evolution_2021}.

Building upon the EvoGym benchmark, recent advancements have significantly improved both the efficiency and effectiveness of the co-optimization process. To navigate vast structural search spaces, probabilistic generative models have gained considerable traction. For instance, continuous latent representations of voxel designs can now be learned via variational autoencoders like MorphVAE \cite{song_morphvae_2024}, and classifier-guided diffusion models have proven highly effective in generating high-fitness morphologies \cite{liu_morphology_2025}. Beyond traditional generative models, search paradigms have further expanded to incorporate Large Language Models as intelligent heuristic operators, exemplified by the LASeR framework \cite{song_laser_2024}. To mitigate the computational cost of repeatedly training controllers from scratch, current methodologies place a strong emphasis on knowledge transfer and inheritance mechanisms. Notable strategies within this vein include curriculum-based approaches that progressively scale structures from small to large topologies \cite{wang_curriculum-based_2022}, the deployment of Transformer-based universal controllers to enable cross-task morphological knowledge transfer \cite{zhao_cross-task_2025}, and Lamarckian transfer learning frameworks that allow for the direct inheritance of neural network weights across generations \cite{harada_lamarckian_2024}. Complementing these transfer methodologies, pre-training paradigms have also been effectively leveraged; for example, brain-body pre-training designs now facilitate few-shot generalization across unseen environments \cite{wang_preco_2023}. Beyond training efficiency, optimizations at the evaluation level have significantly accelerated the design process through surrogate models based on action inheritance \cite{liu_rapidly_2024}. Furthermore, in the context of complex modular systems, multi-objective optimization frameworks have successfully guided the transition of control strategies from simple isolated modules to serially connected tasks \cite{zhao_multi-objective_2025}. Despite these remarkable advancements, none of the aforementioned studies have considered the material dimension as a first-class design variable.

\subsection{Policy Optimization in Deep Reinforcement Learning}

Deep reinforcement learning provides a critical framework for evaluating soft robotic systems, successfully scaling traditional reward-maximizing processes to handle their inherently high-dimensional, non-linear state spaces\cite{mnih_human-level_2015, sutton1998reinforcement}. To mitigate training instabilities in these complex continuous domains, contemporary approaches prioritize robust policy gradient methods. Most notably, Proximal Policy Optimization (PPO) utilizes a clipped surrogate objective to safely constrain policy updates\cite{schulman_proximal_2017}. These frameworks effectively facilitate the assessment and deployment of reactive, closed-loop control in compliant structures\cite{haarnoja2018soft}.

\section{EvoGymCM Benchmark Suite}

To systematically address the limitations of discrete material representations, we propose the EvoGymCM benchmark suite. This section formally defines the underlying physical mechanics and introduces the two benchmark settings tailored for distinct real-world material properties.

\subsection{Robot Representation}

In the EvoGymCM benchmark suite, we fundamentally augment the conventional voxel-based soft robot representation by introducing a continuous material dimension. Physically, the robot executes tasks as the controller drives the rhythmic expansion and contraction of actuated voxels, which in turn induces passive deformation in the surrounding structural voxels. By making the stiffness of these structural elements continuously tunable, our framework breaks the native limitation of strictly fixed materials, transforming compliance from a predefined passive parameter into a highly dynamic design space.

\begin{figure}[tbp] 
    \centering    \includegraphics[width=1.0\columnwidth]{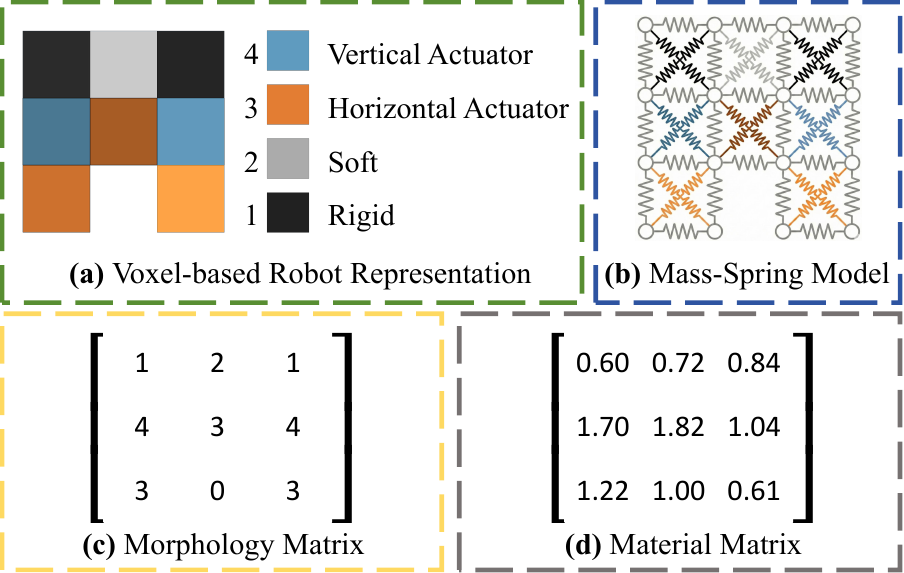} 
    \caption{Robot Representation and Physical Modeling}
    \label{fig:robot}
\end{figure}

To systematically formalize this upgraded design space, we establish a 2D grid workspace with spatial dimensions $W \times H$. Within this space, we formulate the complete co-design problem into three orthogonal dimensions:

\begin{itemize}
    \item \textbf{Morphology ($M$):} Formulated as a discrete matrix $M \in \{0, 1, 2, 3, 4\}^{W \times H}$, this dimension strictly dictates the macroscopic spatial distribution of voxel types. As illustrated in Fig.~\ref{fig:robot}(a) (c), it categorizes each body voxel into one of four distinct types: rigid, soft, horizontal actuator, or vertical actuator.
    
    \item \textbf{Material Stiffness ($S$):} Formulated as a continuous matrix $S \in [0.5, 2]^{W \times H}$ (Fig.~\ref{fig:robot}(d)), this newly introduced dimension governs physical compliance. Rather than encoding absolute stiffness, $S$ acts as a relative scaling field. It assigns an independently searchable, continuous multiplier to the baseline stiffness of the structural voxels defined in $M$.
    
    \item \textbf{Control ($C$):} Parameterized as a learned neural policy, this dimension drives the robot's dynamic behavior. At each time step, $C$ maps real-time state observations to target expansion or contraction ratios for actuators.
\end{itemize}

\begin{figure*}[t] 
    \centering    \includegraphics[width=1.0\textwidth]{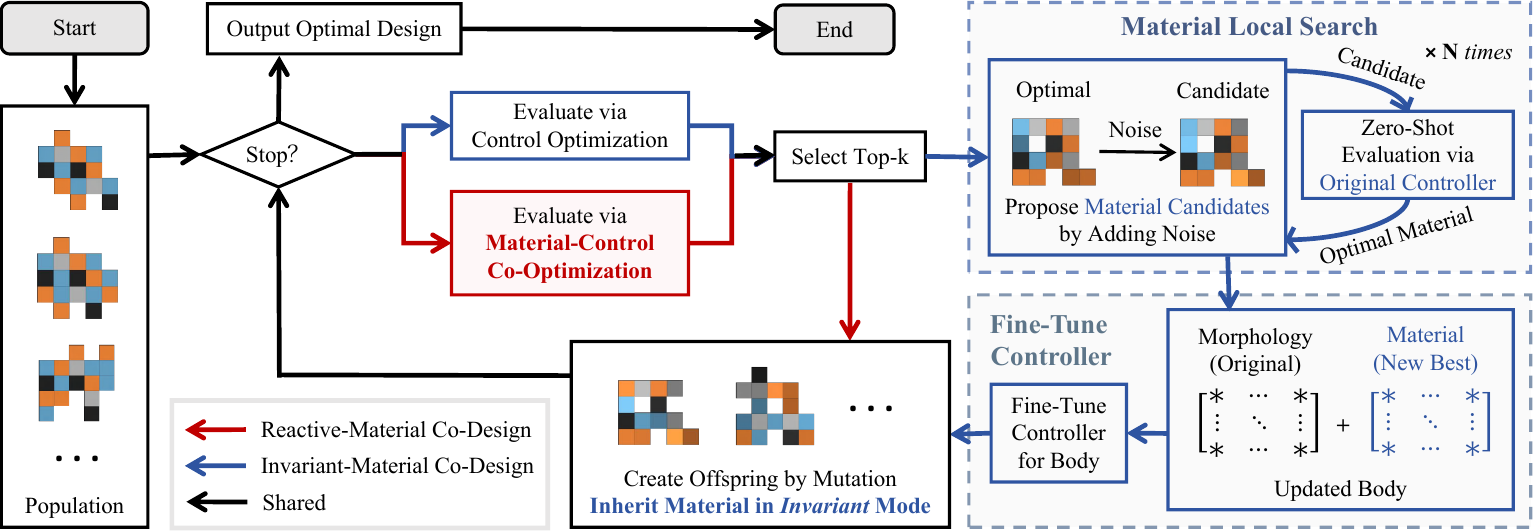} 
    \caption{Algorithmic pipeline of the proposed morphology-material-control co-design paradigms, illustrating the distinct workflows for the Reactive-Material Co-Design (red arrows) and Invariant-Material Co-Design (blue arrows), alongside shared optimization processes (black arrows).}
    \label{fig:method}
\end{figure*}

\subsection{Continuous Material Stiffness Dimension}

The backend dynamics of the environment are governed by a 2D mass-spring physics engine. As illustrated in Fig.~\ref{fig:robot}(b), the structural integrity of each voxel is maintained by an arrangement of point masses and elastic springs. Geometrically, each voxel is modeled as a cross-braced square, with its edges and diagonals acting as ideal springs that strictly obey Hooke's law. In the standard EvoGym implementation, the baseline spring constant, denoted as $k_{base}$, is rigidly locked to a predefined discrete value dictated entirely by the voxel's categorical type in the morphology matrix $M$.

To physically realize the newly introduced continuous material stiffness dimension, EvoGymCM dynamically maps the relative scaling matrix $S$ directly to the underlying spring constants. For an occupied voxel at spatial coordinates $(i, j)$ with a continuous stiffness factor $S_{i, j}$, the scaling is strictly local for internal diagonal springs, whereas boundary springs shared with an adjacent voxel at $(i', j')$ adopt the average of their respective factors:
\begin{equation}
    k_{\text{effective}} = 
    \begin{cases} 
    S_{i,j} \cdot k_{\text{base}}, & \text{diagonal spring} \\
    \frac{1}{2}(S_{i,j} + S_{i',j'}) \cdot k_{\text{base}}, & \text{edge spring}
    \end{cases}
\end{equation}

By deeply embedding this continuous multiplier into the foundational dynamics, we successfully transition soft robot design from discrete combinatorial selection into searchable continuous field optimization.

\subsection{EvoGymCM-R: Benchmark for Programmable Materials}

To physically align with smart programmable materials, we introduce \textit{EvoGymCM-R (Reactive)}. This setting empowers soft robots to dynamically react to their surroundings. By formulating the stiffness field $S_t$ as a time-variant state, the robot can sense environmental feedback and reactively modulate its physical compliance in real-time during environmental interactions.

\textbf{Underlying Physics Implementation:} Rather than permanently overwriting the inherent baseline spring constants, EvoGymCM-R realizes this dynamic responsiveness by passing a real-time material matrix $S_t$ directly into the force calculation phase at each simulation step. Specifically, before computing the elastic forces, the effective spring constant for each voxel is transiently instantiated by multiplying its baseline constant by the corresponding real-time scaling factor from $S_t$. This step-wise formulation precisely replicates the physical mechanics of reactive stiffness tuning, while ensuring absolute numerical stability within the mass-spring solver.

\textbf{Augmented Action Space:} To enable the intelligent regulation of this reactive property, we integrate the material scaling factors directly into the Reinforcement Learning (RL) action space. In this formulation, the control policy outputs an augmented action vector $\mathbf{a}_t = [\mathbf{c}_t, \mathbf{s}_t]$ at each time step, where $\mathbf{c}_t$ represents the traditional motor actuation signals and $\mathbf{s}_t$ denotes the continuous stiffness modulation commands for the material matrix $S_t$. By coupling motor control with reactive material adaptation, the agent learns sophisticated strategies to actively exploit variable compliance on the fly.

Ultimately, by integrating the real-time material matrix into the force calculation phase, we successfully realize programmable materials capable of reactive stiffness modulation during environmental interactions.

\subsection{EvoGymCM-I: Benchmark for Traditional Materials}

To physically align with traditional material manufacturing, where physical properties must be fully determined prior to deployment, we introduce \textit{EvoGymCM-I (Invariant)}. In this setting, the continuous stiffness field $S$ serves as a highly searchable variable exclusively during the offline design phase. Once deployed, $S$ becomes strictly time-invariant, and the robot leverages this pre-optimized static stiffness distribution for subsequent performance evaluation.

\textbf{Underlying Physics Implementation:} In contrast to the step-wise recalculation in EvoGymCM-R, EvoGymCM-I physically realizes the material properties solely during the initialization phase. The optimized continuous stiffness matrix $S$ serves as a spatial mapping, directly scaling the baseline spring constants $k_{\text{base}}$ to define the effective spring constants $k_{\text{effective}}$ for all occupied voxels. Once instantiated, this highly heterogeneous stiffness distribution remains fixed throughout the simulation, fundamentally governing the inherent compliance and force transmission pathways of the soft robot.

Ultimately, by integrating the time-invariant material matrix strictly during the initialization phase, we successfully establish material stiffness as an independently optimizable dimension within the design space.

\section{Morphology-Material-Control Co-Design Algorithmic Paradigms}

To tackle the highly coupled morphology--material--control triad denoted as $\langle M, S, C \rangle$, we navigate this high-dimensional optimization challenge by proposing two distinct co-design algorithmic paradigms. These paradigms are specifically tailored to the \textit{EvoGymCM-R} and \textit{EvoGymCM-I} benchmarks.

\subsection{Reactive-Material Co-Design (Under EvoGymCM-R)}

In the context of smart programmable materials, the dynamic material stiffness field $S_t$ is inherently coupled with real-time motion control $C_t$. At each time step $t$, the learned policy $\pi_\theta$ outputs an augmented action vector $\mathbf{a}_t = [\mathbf{c}_t, \mathbf{s}_t]$. Consequently, the optimization objective is to discover the optimal morphology $M^*$ alongside a policy $\pi_{\theta^*}$ that maximizes the expected return by jointly controlling motor actuation and material stiffness:
\begin{equation}
    \max_{M} \max_{\theta} \mathbb{E}_{\tau \sim \pi_{\theta}} \left[ \sum_{t=0}^T r(\mathbf{c}_t, \mathbf{s}_t \mid M) \right]
\end{equation}

Fig.~\ref{fig:method} details Reactive-Material Co-Design paradigm. The framework operates through a nested bi-level optimization process. The outer loop is dedicated to morphological evolution. For every proposed morphology $M$, the inner loop employs Material-Control Co-Optimization. Here, a reinforcement learning agent is trained to jointly execute motor control and real-time stiffness tuning for the given body, enabling the soft robot to dynamically react to environmental interactions. By alternating between these two loops, the algorithm efficiently identifies the optimal morphology and policy pair $\langle M^*, \pi_{\theta^*} \rangle$. 

Ultimately, the resulting policy $\pi_{\theta^*}$ serves as an adaptive modulation policy, guiding programmable materials to reactively regulate their stiffness in real-world applications.

\subsection{Invariant-Material Co-Design (Under EvoGymCM-I)}

In the context of traditional material manufacturing, the material stiffness field $S$ and morphology $M$ jointly constitute a time-invariant physical body $B = \langle M, S \rangle$. Unlike the reactive setting, $S$ here is a fixed structural parameter. Consequently, the control policy $\pi_\theta$ solely outputs motor actuation commands $\mathbf{c}_t$. The objective is to co-optimize this unified physical body alongside its corresponding control policy:
\begin{equation}
    \max_{M, S} \max_{\theta} \mathbb{E}_{\tau \sim \pi_{\theta}} \left[ \sum_{t=0}^T r(\mathbf{c}_t \mid M, S) \right]
\end{equation}

Fig.~\ref{fig:method} details Invariant-Material Co-Design paradigm. In contrast to the reactive approach where material adaptation occurs within the inner RL loop, this paradigm elevates material optimization to the outer loop. 

This approach prioritizes structural evolution while utilizing a \textit{Material Local Search} and a \textit{Fine-Tune Controller} phase to efficiently optimize the material distribution $S$ within a fixed morphology.
Specifically, the \textit{Material Local Search} exclusively targets the top-performing survivors. For each surviving morphology, the algorithm iteratively adds noise to the original material matrix to generate new candidates. These candidates then undergo zero-shot evaluations via the original controller. If a candidate achieves a higher reward, it immediately overwrites the current material matrix. 
After this iterative search concludes, the \textit{Fine-Tune Controller} phase updates the inherited policy for this newly assembled embodiment directly. 

Through this decoupled process, the algorithm systematically outputs the globally optimal physical body $B =  \langle M^*, S^* \rangle$ and its dedicated policy $\pi_{\theta^*}$. 

Ultimately, the optimized material stiffness distribution $S$ provides quantitative guidance for physical fabrication (e.g., multi-material 3D printing infill densities and spatial mixing ratios), facilitating the realization of traditional material manufacturing.

\section{Experiments and Results}

To comprehensively evaluate the impact of introducing a continuous material stiffness dimension, we conducted systematic experiments across diverse task environments. Beyond quantifying pure performance improvements, our analysis deeply investigates the underlying mechanisms: specifically, the synergistic coupling between material properties and the control policy, and how this expanded design space influences morphological evolution.

\subsection{Experimental Setup}

To verify the generalizability of our proposed algorithms, we selected six representative tasks, categorizing them by the specific physical challenges they impose on the soft robots:

\begin{itemize}

    \item \textbf{Walking:} \texttt{Walker-v0} and \texttt{BridgeWalker-v0}. These tasks evaluate the robot's fundamental forward locomotion capabilities on relatively flat or gapped terrains.
    \item \textbf{Locomotion:} \texttt{DownStepper-v0} and \texttt{CaveCrawler -v0}. These tasks evaluate the robot's physical compliance and obstacle negotiation skills in constrained spaces and rugged, uneven topographies.
    \item \textbf{Shape Change:} \texttt{AreaMaximizer-v0} and \texttt{AreaMinimizer-v0}. These tasks evaluate the robot's extreme deformation capabilities by requiring the maximization and minimization of its spatial occupancy.

\end{itemize}

\subsection{Evaluation of Reactive-Material Co-Design}

\noindent \textbf{1) Material-Control Co-Optimization vs. Control-Only} 

\begin{figure}[tbp]
    \centering    \includegraphics[width=1.0\linewidth]{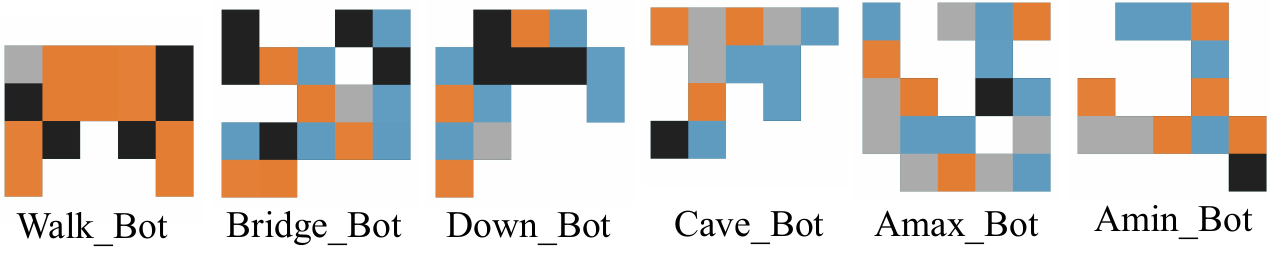}
    \caption{Fixed Robot Set}
    \label{fig:diffrobot}
\end{figure}

To isolate the interference of outer-loop morphological evolution, we conducted an inner-loop ablation using fixed, pre-designed morphologies ($M$) (as shown in Fig.~\ref{fig:diffrobot}). We compared our proposed Material-Control Co-Optimization against conventional Control Optimization, which operates under fixed material properties and generates purely kinematic control signals. Table~\ref{tab:comparison} summarizes the quantitative results averaged across three random seeds.

\begin{table}[h]
\centering
\caption{Performance Comparison Across Different Tasks} 
\label{tab:comparison}
\renewcommand{\arraystretch}{1.3} 
\begin{tabular}{l l c c}
\toprule
\textbf{Task} & \textbf{Robot} & \textbf{Control-only} & \textbf{Co-Opt} \\
\midrule
Walker-v0       & Walk\_Bot   & 9.46 (0.03) & \textbf{9.55 (0.01)} \\
BridgeWalker-v0 & Bridge\_Bot & 2.10 (0.23) & \textbf{2.97 (0.24)} \\
DownStepper-v0  & Down\_Bot   & 3.25 (0.27) & \textbf{3.92 (0.16)} \\
CaveCrawler-v0  & Cave\_Bot   & 2.02 (0.01) & 2.03 (0.01) \\
AreaMaximizer-v0& Amax\_Bot   & 0.62 (0.07) & \textbf{0.74 (0.14)} \\
AreaMinimizer-v0& Amin\_Bot   & 0.61 (0.04) & \textbf{0.69 (0.04)} \\
\bottomrule
\end{tabular}
\end{table}

As shown in Table~\ref{tab:comparison}, introducing the continuous material dimension yields performance improvements closely tied to environmental complexity. In fundamental walking tasks (\texttt{Walker-v0}), gains are marginal as flat terrains impose minimal compliance demands, allowing the counterpart to perform near optimality. Conversely, in complex environments (\texttt{BridgeWalker-v0}, \texttt{DownStepper-v0}) and extreme shape-change tasks (\texttt{AreaMaximizer-v0}, \texttt{AreaMinimizer-v0}), Co-Adaptation achieves substantial score increases, demonstrating reactive material tuning significantly enhanced terrain adaptability and structural deformability. Interestingly, \texttt{CaveCrawler-v0} shows negligible improvement due to a hard physical bottleneck: the narrow gap requires a morphological volume reduction that stiffness modulation alone cannot achieve.

\begin{figure}[tbp]
    \centering
    \includegraphics[width=1\linewidth]{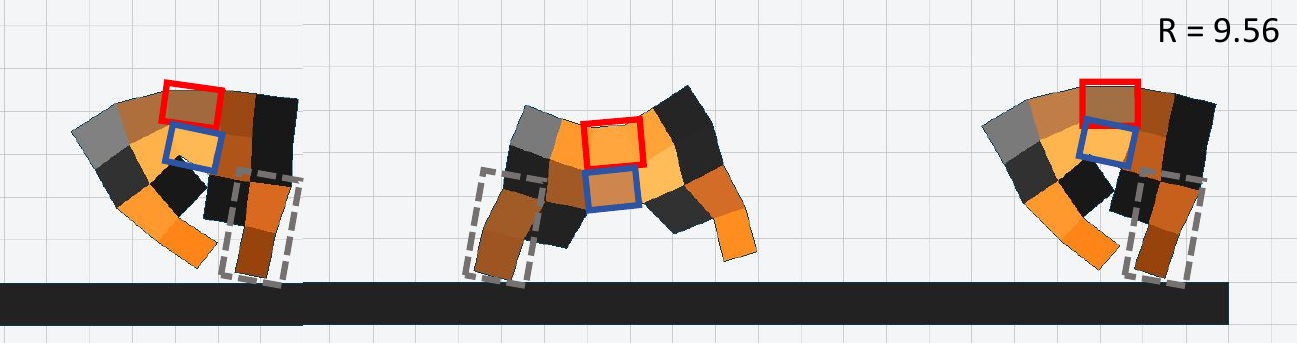}
    \caption{Representative Case of Material-Control Co-Optimization}
    \label{fig:walk}
\end{figure}

\textbf{Reactive Behavioral Analysis.}
Beyond numerical gains, visual analysis confirms explicit reactive material behaviors, where the robot dynamically alters its physical properties to overcome environmental challenges. For instance, sequential snapshots in Fig.~\ref{fig:walk} (grey highlighted regions) show the robot autonomously increasing the stiffness of its ``leg'' voxels right before ground impact (darker colors indicate higher stiffness). This anticipatory stiffening ensures structural stability and prevents body collapse under impact forces. 

\begin{figure}[tbp]
    \centering
    \includegraphics[width=1\linewidth]{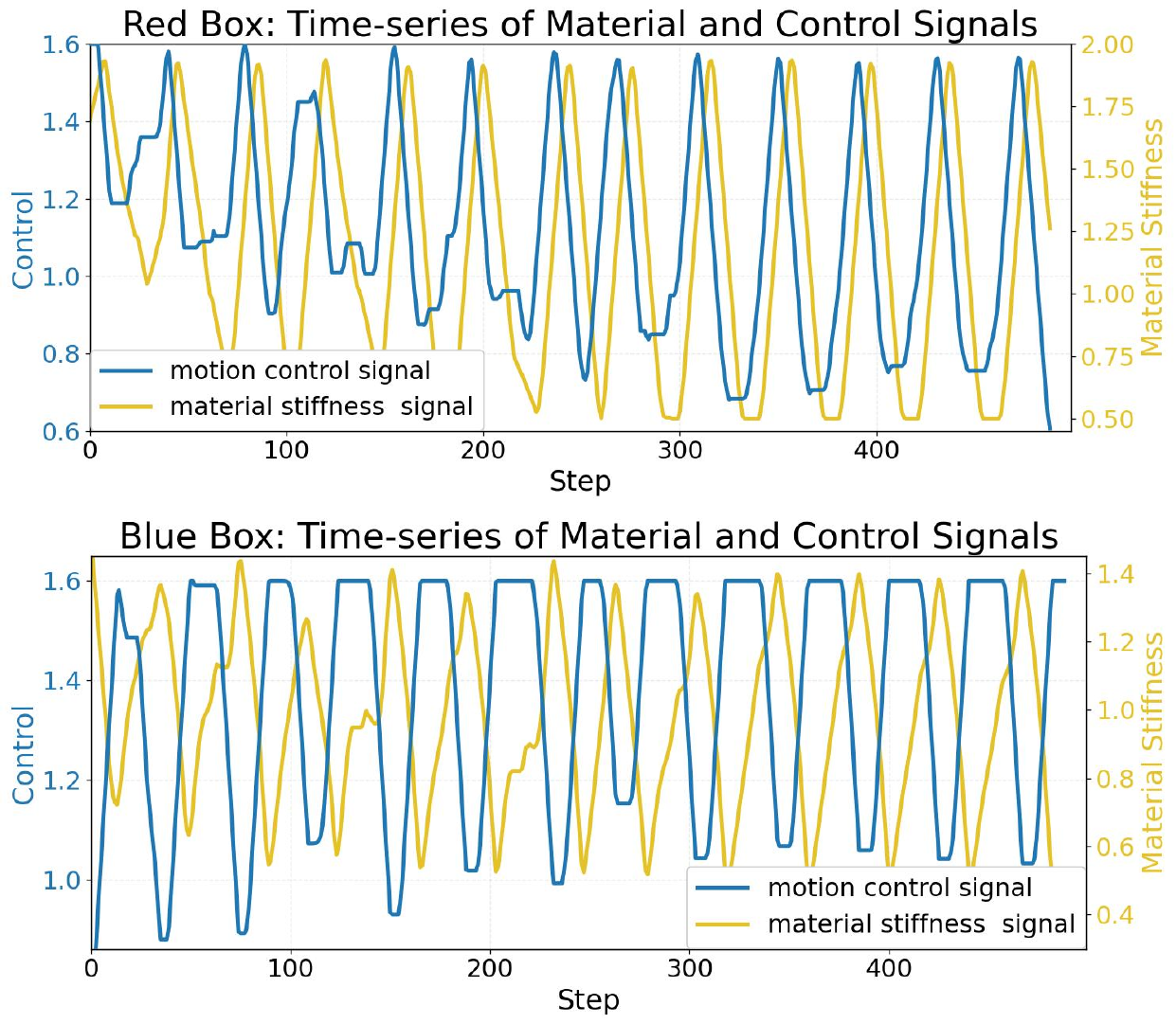}
    \caption{Correlation between Material Stiffness and Control Signals}
    \label{fig:signal}
\end{figure}

\textbf{Quantitative Analysis of Reactive Synergy.} 
To uncover the underlying mechanism of this synergy, we analyzed the real-time signal outputs of two representative voxels (highlighted in red and blue boxes in Fig.~\ref{fig:walk}), revealing a strong statistical correlation between motor actuation ($\mathbf{c}_t$) and material stiffness ($\mathbf{s}_t$) signals. As illustrated in Fig.~\ref{fig:signal}, the time-series trajectories of both signals exhibit a strict frequency alignment, confirming a highly coupled cyclical response. While a consistent phase shift is observed between the two curves, this variance fundamentally stems from the inherent \textit{physical actuation lag}---the natural delay between the issuance of a kinematic control command and the resulting morphological deformation. Accounting for this mechanical delay, the precise frequency synchronization explicitly demonstrates their synergistic interplay. Ultimately, this robust coupling verifies that the agent has successfully forged a unified control-material policy, genuinely realizing reactive material adaptation to optimize physical performance.

\noindent \textbf{2) Impact of Reactive-Material on Co-Design}

To evaluate the contribution of real-time stiffness modulation, we conducted a bilevel ablation study focusing on the reactive-material dimension. Specifically, we compared our \textit{Reactive-Material Co-Design} paradigm against a \textit{Fixed-Material} variant. In this setting, the material dimension is restricted to a constant stiffness value throughout the task-environment interactions. This restriction effectively isolates the performance impact of continuous material stiffness tuning, highlighting how reactive adaptation to environmental stimuli serves as the primary driver of superior physical performance.

\begin{figure}[tbp]
    \centering
    \includegraphics[width=1\linewidth]{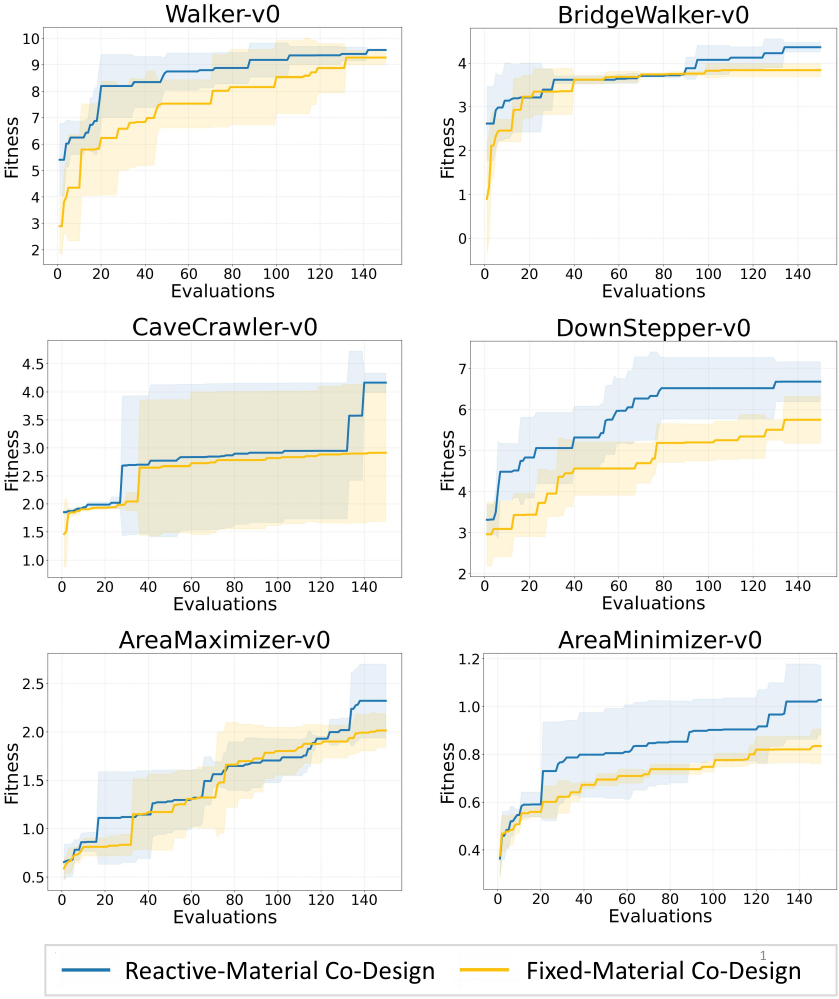}
    \caption{Bilevel ablation results comparing Reactive-Material Co-Design and the Fixed-Material variant across six tasks.}
    \label{fig:react}
\end{figure}

As illustrated in Fig.~\ref{fig:react}, integrating continuous material tuning primarily elevates the performance upper bound of the evolved robots rather than altering the base efficiency of the morphological search. Because both methods share the same robust outer-loop exploration strategy, their initial convergence rates are naturally comparable; however, the inner-loop reactive co-adaptation fundamentally unlocks a higher physical potential for these morphologies, significantly raising the task-specific performance ceiling. This enhancement manifests differently depending on the tasks:
\begin{itemize}
    \item \textbf{Walking:} For \texttt{Walker-v0} and \texttt{BridgeWalker-v0} tasks, the performance gain is relatively modest, as basic walking relies more on overall stride scale rather than extreme structural deformation.
    \item \textbf{Locomotion:} For \texttt{CaveCrawler-v0} and \texttt{DownStepper-v0} tasks, enhanced deformability provides crucial terrain adaptability. It allows the agent to navigate narrow gaps and irregular steps, thereby substantially raising the fitness ceiling.
    \item \textbf{Shape Change:} For \texttt{AreaMaximizer-v0} and \texttt{AreaMinimizer-v0} tasks, variable stiffness explicitly enhances physical deformation capabilities, leading to pronounced performance breakthroughs.
\end{itemize}

\begin{figure}[tbp]
    \centering
    \includegraphics[width=0.7\linewidth]{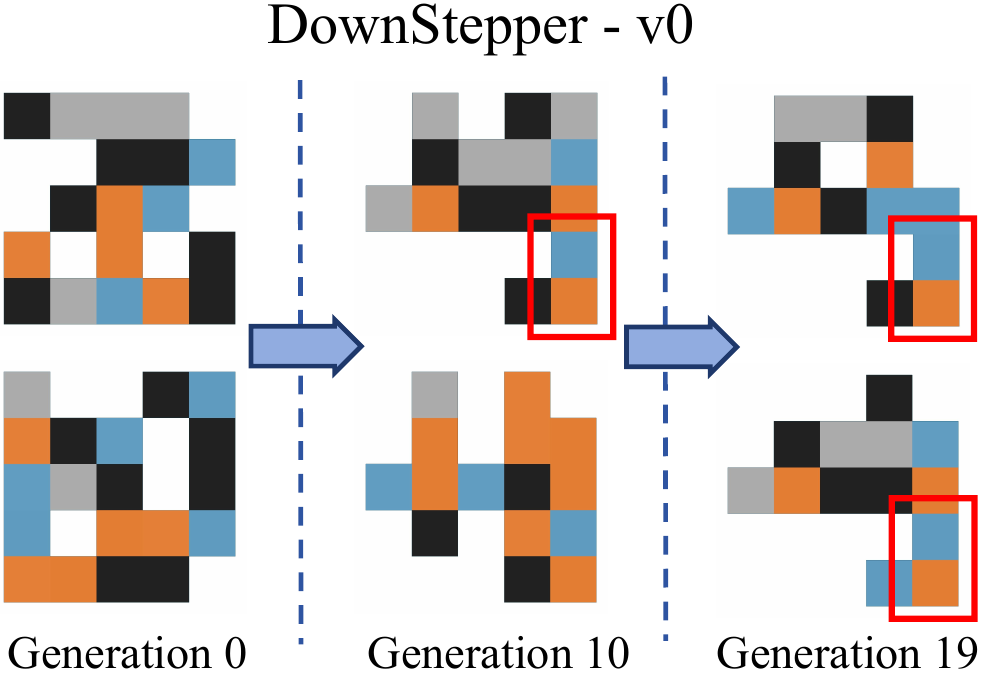}
    \caption{Evolution of Robot Designs via Reactive-Material Co-Design}
    \label{fig:downstep}
\end{figure}

\textbf{The reactive-material dimension reshapes the evolutionary landscape.} Interestingly, we observed that the introduction of reactive materials encourages an evolutionary trend towards structurally narrower and seemingly more fragile morphologies. As illustrated in the evolutionary trajectory of \texttt{DownStepper-v0} (Fig.~\ref{fig:downstep}), the algorithm progressively drives the morphology towards a distinctly thin ``leg'' structure (highlighted by the red box in Fig.~\ref{fig:downstep}). In traditional fixed-material evolution, such a slender limb would be heavily penalized or prematurely eliminated due to its inherent physical instability and vulnerability to impact. However, within our reactive framework, the agent can dynamically regulate its own structural stability by tuning its material stiffness in real-time, safely stabilizing the thin leg to prevent collapse. This dynamic compensation allows the slender morphology to achieve far greater physical compliance and perform exceptionally well in irregular environments, ultimately transforming a traditional structural liability into a highly adaptive morphological advantage.

\subsection{Evaluation of Invariant-Material Co-Design}

\noindent \textbf{Impact of Optimizable Invariant-Material on Co-Design}

To evaluate the effectiveness of invariant-material optimization, we conducted a bilevel ablation study. To isolate the impact of spatial material heterogeneity, we compared our joint optimization approach against a \textit{Prescribed-Material Co-Design} variant. Rather than using a complex joint search space ($\{\mathbf{M}, \mathbf{S}\}$), this restricted variant optimizes solely for the discrete morphological topology ($\mathbf{M}$) while locking the material dimension to a hard-coded, prescribed stiffness value throughout the evolutionary process.

\begin{figure}[tbp]
    \centering
    \includegraphics[width=1\linewidth]{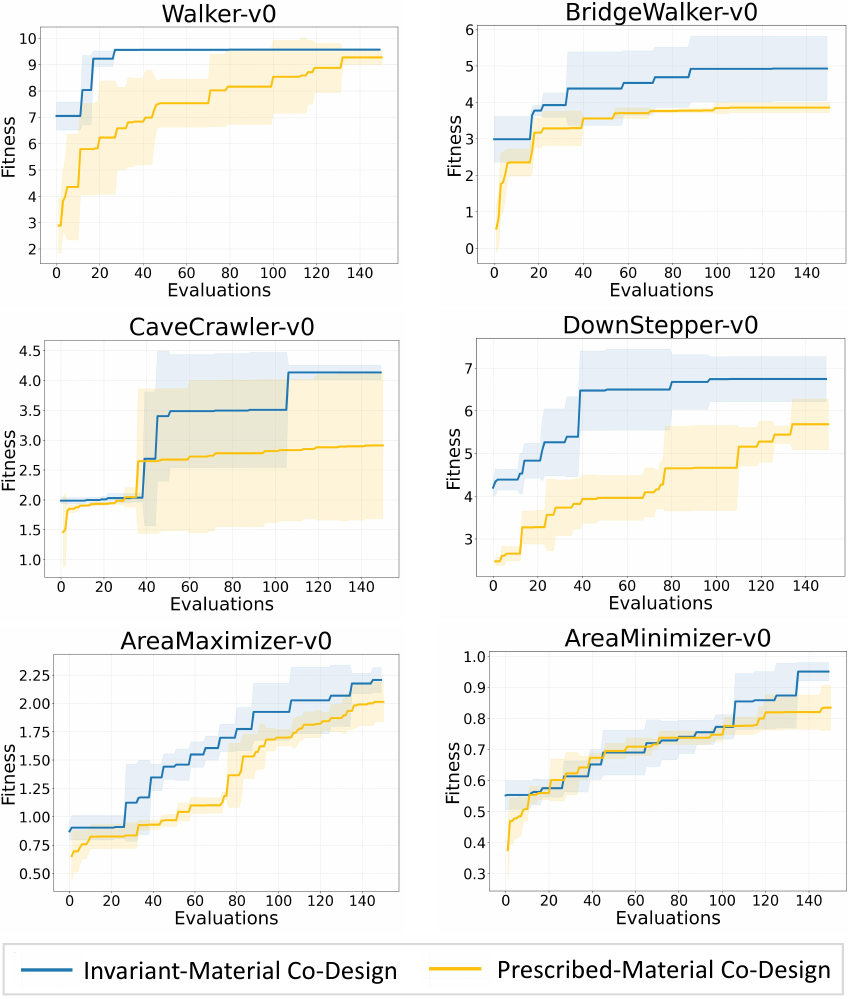}
    \caption{Bilevel Ablation Results comparing Invariant-Material Co-Design and the Prescribed-Material variant across six tasks.}
    \label{fig:invariant}
\end{figure}

As illustrated in Fig.~\ref{fig:invariant}, integrating the invariant material dimension simultaneously accelerates evolutionary efficiency and elevates the performance upper bound. Specifically, across the two walking tasks (\texttt{Walker-v0} and \texttt{BridgeWalker-v0}) and the two locomotion tasks (\texttt{CaveCrawler-v0} and \texttt{DownStepper-v0}), we observe substantial improvements in both convergence speed and terminal fitness. This demonstrates that discovering an optimal spatial material distribution effectively unlocks the physical potential of the robots, validating that our \textit{Invariant-Material Co-Design} paradigm successfully identifies synergistic material fields that complement the evolved morphologies. Conversely, for the two shape-changing tasks (\texttt{AreaMaximizer-v0} and \texttt{AreaMinimizer-v0}), the performance gains are relatively marginal. This plateau likely occurs because achieving extreme global deformation requires highly precise, voxel-specific stiffness tuning. The current evolutionary paradigm, which optimizes a global spatial field, may lack the micro-level precision needed to isolate and drastically alter just a few critical voxels, rendering the performance improvements less pronounced in extreme morphing scenarios.

\begin{figure}[tbp]
    \centering
    \includegraphics[width=0.7\linewidth]{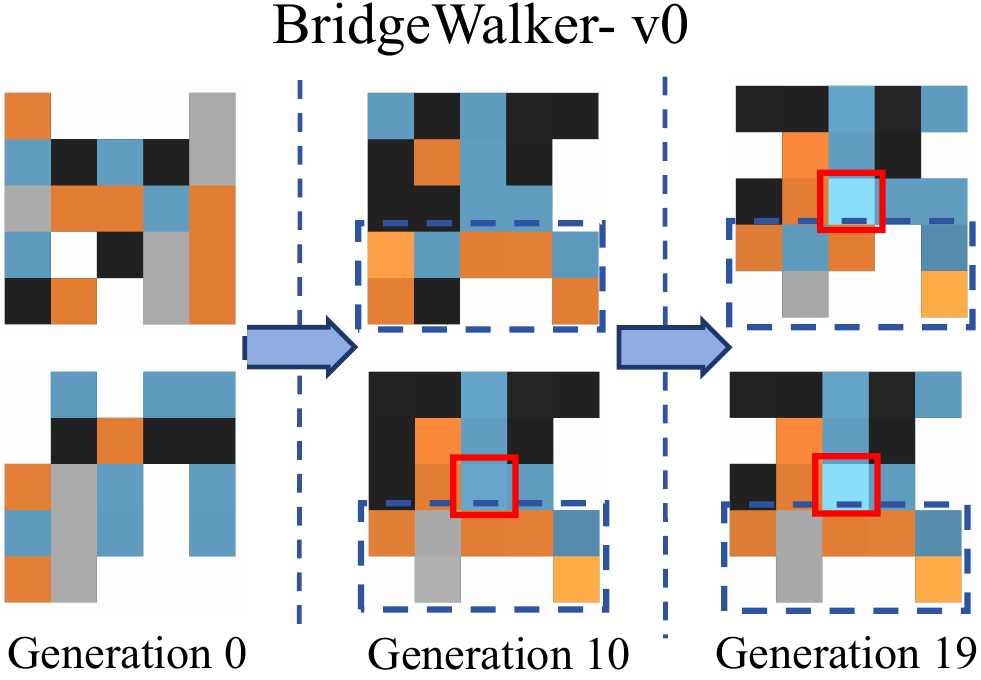}
    \caption{Evolution of Robot Designs via Invariant-Material Co-Design}
    \label{fig:bridge}
\end{figure}

\textbf{Coupled Regularities in Material-Morphology Synergy.} Rather than drastically altering the macroscopic shape, the introduction of the spatial material dimension reveals tightly coupled regularities between material distribution and the evolved morphology. Taking the task \texttt{BridgeWalker-v0} as an illustrative example (Fig.~\ref{fig:bridge}), two distinct, synergistic material patterns emerge that directly contribute to task success. First, the central body voxel (highlighted by the red box in Fig.~\ref{fig:bridge}) exhibits a clear evolutionary trend toward lower stiffness. This localized softening minimizes internal mechanical constraints, granting the body superior deformability for complex maneuvers. Second, the robot consistently evolves a high-stiffness base (blue dashed box in Fig.~\ref{fig:bridge}) that acts as a structural ``chassis'', significantly improving locomotion stability. Ultimately, these purposeful material-morphology couplings prove that optimizing spatial material fields is crucial for unlocking the full physical capabilities of soft robots.

\section{Conclusion and Future Work}

In conclusion, this paper introduces the EvoGymCM benchmark suite, formally establishing continuous material stiffness as a first-class design variable and proposed two settings---\textit{R} and \textit{I}---accompanied by the Reactive-Material and Invariant-Material co-design paradigms. Systematic experiments demonstrate that integrating the continuous material dimension elevates the performance ceiling of soft robots through deep synergy with morphology and control. 

For future work, we plan to develop more sample-efficient algorithms to navigate this expanded, high-dimensional design space. Furthermore, we aim to bridge the Sim-to-Real gap, deploying the optimally evolved designs and policies onto physical robotic platforms.

\end{document}